# Déjà vu: Scalable Place Recognition Using Mutually Supportive Feature Frequencies

Adam Jacobson, Walter Scheirer and Michael Milford, *Member IEEE*

*Abstract*— Learning and recognition is a fundamental process performed in many robot operations such as mapping and localization. The majority of approaches share some common characteristics, such as attempting to extract salient features, landmarks or signatures, and growth in data storage and computational requirements as the size of the environment increases. In biological systems, spatial encoding in the brain is definitively known to be performed using a fixed-size neural encoding framework - the place, head-direction and grid cells found in the mammalian hippocampus and entorhinal cortex. Particularly paradoxically, one of the main encoding centers - the grid cells - represents the world using a highly aliased, repetitive encoding structure where one neuron represents an unbounded number of places in the world. Inspired by this system, in this paper we invert the normal approach used in forming mapping and localization algorithms, by developing a novel place recognition algorithm that seeks out and leverages repetitive, mutually complementary landmark frequencies in the world. The combinatorial encoding capacity of multiple different frequencies enables not only the ability to achieve efficient data storage, but also the potential for sub-linear storage growth in a learning and recall system. Using both ground-based and aerial camera datasets, we demonstrate the system finding and utilizing these frequencies to achieve successful place recognition, and discuss how this approach might scale to arbitrarily large global datasets and dimensions.

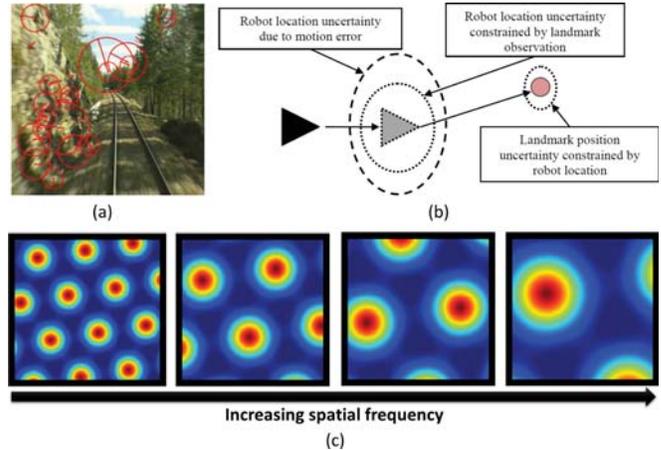

Fig. 1: How robotics and natural systems encode space. (a) Encoding using salient local descriptors and (b) using uniquely identifiable geometric landmarks in a traditional geometric SLAM system. (c) Auto-correlogram of 4 different grid cell firing patterns as a rat moves around in a square arena - showing how each cell encodes a large number of regularly repeating locations in space at a range of scales.

## I. INTRODUCTION

Many robotic and autonomous capabilities including navigation and object, face, voice or scene recognition rely on the storage and processing of large amounts of data against which current sensory input is compared. While the specifics of the type and nature of the data and the way in which it is processed vary from task to task, there are broad shared characteristics. Many of these tasks do not require the original sensory data to stored in a lossless manner; rather they abstract the data to useful signatures or feature-based representations which are sufficient for the specific task. Secondly, regardless of the type of data format, the amount of data that is stored generally scales at least linearly with the size of the environment (in a navigation context) or the number of distinct instances of a thing (in the object, face and voice recognition contexts). While the cost of storage (whether local or in the cloud) is steadily decreasing, this is more than balanced out by the rate at which data is being generated; in fact data generation is generally acknowledged to be outstripping storage capacity [1]. Regardless of whether the technical storage requirements are feasible for a specific task, great value is placed upon using the most efficient storage mechanisms, and much effort is still devoted towards many aspects of data compression [2]. Reducing storage requirements also often has the added benefit of reducing computational requirements - *if* accuracy is comparable, recognizing a face or place using a handful of bytes is generally more attractive than requiring megabytes of raw uncompressed image data.

In this paper, we develop the novel Déjà vu approach for achieving high levels of data compression for any learning and recognition task - a task where reference data is stored in order to be compared against future sensory input, without necessarily requiring the full reconstruction of the original raw data (recall). Many recognition tasks fit this characteristic under a wide range of end user scenarios, including face recognition, object recognition and the domain in which this research lies, place recognition for navigation.

Our approach is inspired by a seeming paradox between how conventional robotic mapping and navigation systems store data about the world and how the mammalian brain encodes space. Robotic navigation systems typically look

A. Jacobson and M. Milford are with the School of Electrical Engineering and Computer Science, Queensland University of Technology, Brisbane, Australia, a1.jacobson@qut.edu.au. MM also with the Australian Centre for Robotic Vision. W. Scheirer is with the Department of Computer Science and Engineering, University of Notre Dame. This work was supported by an Asian Office of Aerospace Research and Development Grant FA2386-16-1-4027 and an ARC Future Fellowship FT140101229 to MM.

for salient, unique cues / features / landmarks [3]–[9] in the environment because they offer the least perceptual aliasing and ambiguity - some of this likely a hangover from when perfect data association was a requirement [3]–[6] (Figure 1(a,b)). Repeating features that are not ignored using saliency measures often introduce the potential for false positive localization results - for example in a road-based domain, highly repetitive features like telegraph poles and road markings. These repetitive features are typically ignored in conventional systems, with only a small amount of work that seeks to use repeating features.

The Déjà vu algorithm flips this convention on its head by actively seeking to identify repetitive features and landmarks in an environment that occur at mutually supportive frequencies. With appropriate selection of frequencies, even two features or landmarks with different repetitive frequencies can enable place recognition across a large section of a route due to their combinatorial encoding capacity. By chunking the environment being encoded into segments that can be fully encoded by mutually supportive landmark frequencies, large datasets can be encoded using a very compact representation.

The paper proceeds as follows. In Section 2, we review robotic and animal mapping systems with a focus on multi-scale representations of the world. Section 3 presents our approach describing in detail the place recognition system and the proposed multi-scale sensor fusion technique. In Section 4, we present the experimental setup, with the results of multiple levels of evaluation presented in Section 5. Section 6 discusses the outcome of the research and areas of future work.

## II. BACKGROUND

In this section, we review relevant mapping and place recognition techniques, memory hashing processes and biological multi-scale grid cell encoding.

Appearance based SLAM systems [3], [4] typically create some type of database for the environment they are mapping. As a robot explores a new environment, the database grows along with the associated computational and storage requirements. Many techniques reduce the growth in computational requirements by using highly efficient matching and mapping algorithms [10], although these approaches only ameliorate rather than remove the underlying storage growth problem. Other techniques involve some form of map maintenance or pruning [11], which typically reduce the effective resolution of the stored map based on information theoretic principles or heuristics.

The problem of encoding arbitrary-sized datasets with finite storage and compute is highly-relevant in machine learning. A hash function is a deterministic function, $f(x)$, which transforms large data, $x$, into a more compact representation such as a hash table. Hash functions are often used to speed up database lookups and are also used in file checksums, due to the difficulty of inverting the function. An ideal hash function can uniquely represent all possible values in the input dataset, $x$. This unique mapping is often not possible because of computational or storage constraints. Instead, typical hashing approaches introduce the possibility of hash collisions, where $f(x) = f(y)$ and $x \neq y$. A number of techniques for avoiding or dealing with hash collisions have been proposed in the literature.

It appears likely that mammals such as rodents employ some form of hash-like representation of space, using a type of neuron in the rat brain called a grid cell that encodes multiple places simultaneously. Theoretical studies have shown that the multiple map scales [12], [13] in the rodent brain may enable a rodent to uniquely disambiguate a place even when each network in isolation is overloaded and incapable of doing so. Theoretically, an arbitrarily large environment can be encoded using these parallel mapping scales, depending on the degree of noise rejection and error tolerance that is required [12], [13].

Mathematical studies of grid cells in the mammalian brain have shown that very large environments can, at least theoretically, be encoded by a relatively small number of grid cells with different spatial frequencies. These studies generally only assess scenarios with noisy but drift-free self-motion signals, and do not tackle the problem of data association which is covered here.

Based on the precedent set by nature, in this paper we investigate whether correct global location estimates can be extracted by combining location estimates from repetitive features in the environment that repeat at different frequencies.

## III. APPROACH

In this section, we outline the approach of the Déjà vu algorithm, highlighting the compression stages implemented within the database learning phase and the generation of unique global place estimates within the place recognition stage.

### A. Database Learning

The database learning approach within Déjà vu requires the selection and learning of repeating features within an environment. We use an abstraction of the grid cell formation found within the rodent hippocampus, seen in Figure 1. We create multiple "neural" layers consisting of grid cells which encode locations within the environment. Each neural layer encodes features which are active at a particular frequency within the environment. Individual neurons within each neural layer represent the phase of each periodic feature and store an exemplar template to enable future localization.

Periodic features are found within databases by performing a spectrogram analysis for each feature through time. The spectrogram analysis identifies local areas, or chunks, within the database exhibiting periodic activations. Features with similar periodicity, within similar regions or chunks within the database are clustered together and for each phase offset within the periodic region, exemplar templates are created. We only utilize the two strongest frequency results for each feature, as determined by the signal power measured by the spectrogram. For a set of features, with a period of $\tau$, $\tau$ exemplar templates are created. The exemplar templates are created by finding the mean of the features with the same phase offset. The mean $\mu$ for each phase offset $k$, period $\tau$ and feature $H$ is then determined using:

$$\mu_{(H,\tau,k)} = \frac{\sum_{i=0}^{n} H(i) M(i,\tau,k)}{\sum_{i=0}^{n} M(i,\tau,k)} \quad (1)$$

where $i$ is the database frame number, $n$ is the total number of training frames, $H(i)$ is feature component for frame number $i$ and $M$ is the temporal activation of each feature at a particular phase offset calculated using:

$$M(i,\tau,k) = \begin{cases} 1, & \text{if } mod(i-k,\tau) == 0 \\ 0, & \text{otherwise} \end{cases} \quad (2)$$

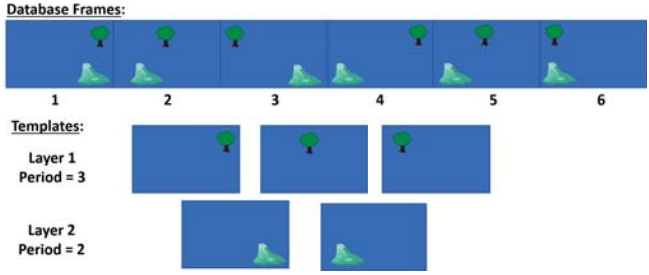

Fig. 2: Illustration of the Déjà vu algorithm, demonstrating how the environment is encoded for compression and later place recognition. Each neural layer represents features which are periodic within an environment. Repeating features within the database are identified and used to compress the database. Each neural layer represents features which are periodic within an environment. Each "neuron" or template within each neural layer encodes the phase offset within the repeating feature pattern and stores an exemplar template to enable future localization.

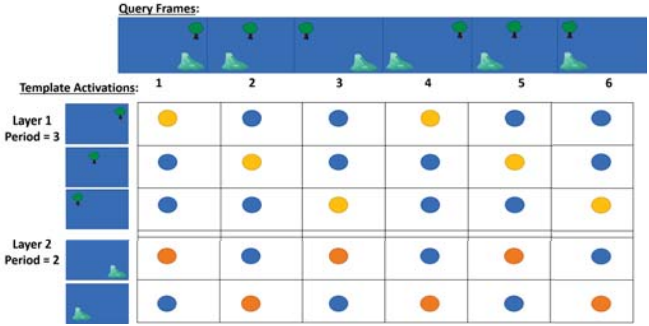

Fig. 3: Illustration of the template activations as query frames change, notice the multiple activations of individual features at multiple query frames. Utilizing a single layer would not enable localization, however multiple layers enable unique localization.

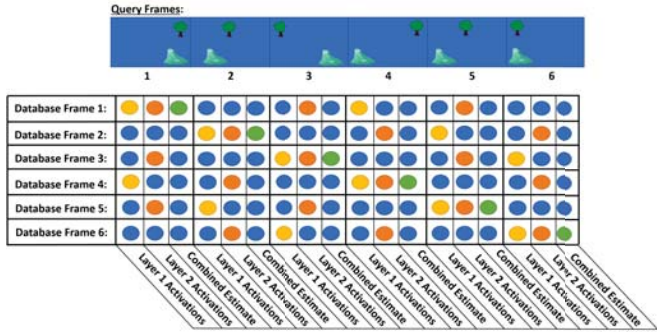

Fig. 4: Unique localisation is achieved through the unique co-occurrence of repeating features.

As a result, the Compressed Database is represented for each layer $C \in R^{F \times P}$ where F is the number of features within the frequency cluster, where $F <= m$, where $m$ is the number of features per frame and $P$ is the number of integer phase offsets within a frequency cluster. For a given dataset, containing repetitious features, this compressed representation would enable significant reduction in required storage, compared to the default requirement of storing features for each individual frame: $T \in R^{n \times m}$. We note that for a particular set of frequency clusters, so long as the periods, $\tau$, for the frequency clusters are co-prime, the compressed representation can represent a total number of locations equal to the product of the period of the frequency clusters ($\prod(\tau)$), whilst only needing to store a maximum number of templates equal to the sum of the period of the frequency clusters ($\sum(\tau)$).

Once templates have been learned for each phase offset of each frequency layer, the compressed representation can be utilized for localization.

### B. Place Recognition

Performing localization using the compressed database representation involves comparing the feature representation of the current input query image to the compressed feature representation using the an $L_1$ norm (however the specific comparison is not important and this could be changed to another metric). For each phase offset, $k$, for each dominant period $\tau$, the matching score, $s_{\tau,k}(i)$, is defined as:

$$s_{\tau,k}(i) = \begin{cases} \sum_{H=1}^{F} |H(i) - \mu_{(H,\tau,k)}| & \text{if } mod(i-k,\tau) == 0 \\ 0, & \text{otherwise} \end{cases} \quad (3)$$

The combined matching score, $S(i)$, is defined as the combination of matching scores from the set of all frequency layers, $L$, using:

$$S(i) = \sum_{\tau \in L} \sum_{k=1}^{\tau} s_{\tau,k}(i) \quad (4)$$

### C. Selecting a Place Match

To determine if a query location matches a reference location, a search is performed for the place hypothesis with the smallest combined sensor matching score:

$$b = \operatorname*{argmin}_{i \in D}(S(i)) \quad (5)$$

where $b$ is the best place hypothesis for the current query image and $D$ is the set of all locations in the database traverse.

The current best matching sensory snapshot is determined to be a place match if the difference score is below a global matching threshold $s_{thresh}$:

$$m = \begin{cases} 1, S(b) \geq s_{thresh} \\ 0, S(b) < s_{thresh} \end{cases} \quad (6)$$

It is this threshold, $s_{thresh}$, which determines if a particular location is a place match.

## IV. EXPERIMENTAL SETUP

In this section, we describe the datasets, testing environment and comparison metrics utilized for evaluating the proposed approach. Datasets for the test environments are available for readers to download at the following link: https://wiki.qut.edu.au/display/cyphy/Datasets

### A. Datasets

To evaluate the proposed method, we utilized three different datasets; the Overhead, the Outdoor and the Underground datasets. These datasets have been selected to illustrate the performance of the proposed Déjà vu algorithm in "man-made" environments: we discuss ongoing work in natural environments in the discussion.

*1) Aerial Dataset:* The Aerial dataset was created by leveraging an aerial image of a section of road from Nearmaps. The single aerial image is then processed to create simulated aerial images to represent an aerial vehicle flying overhead, tracking a road way. The dataset has 3072 frames and has two major repeating features throughout the dataset, as seen in Figure 5.

*2) Outdoor Dataset:* The Outdoor dataset is captured using a single camera placed at the front of a train facing toward the railway track. The dataset is a segment of footage captured by the Chicago Transit Authority, approximately 2km long, as a train traverses the "Blue Line" traveling from Forest Park to O'Hare Airport [14]. This segment of dataset consists of an Outdoor region of dataset where the train travels between train stations.

*3) Underground Dataset:* The Underground dataset is captured using an underground portion of the footage used for the Outdoor dataset produced by the Chicago Transit Authority. The dataset captures a train traveling through tunnels and subway stations, approximately 4km long, with strong regions of repetition from such as tunnel lighting present throughout all of the tunnels and structure within the train stations such as roof supports and repeating signage.

### B. Pre-Processing

The pre-processing applied to each image for this approach involved extracting HOG features for each image. In the case of the Outdoor and Underground datasets, we utilize optical flow of the ground directly in front of the train to estimate velocity of the train and velocity normalize the footage to ensure metrically periodic structure in the environment appears periodically within the feature space.

### C. Comparison Metrics

The compressed database representations are compared to the uncompressed database to evaluate the viability of this compression technique in ideal conditions. We evaluate the proposed system using the Precision-Recall metric. Precision is a measure of the number of correct place matches divided by the total number of place matches reported. Recall is measured as the fraction of correct places matches divided by all of the recallable place matches within a dataset. A Precision-Recall curve is a two-dimensional plot with the X-axis indicating recall and the Y-axis indicating precision. Measuring the maximum recall at 100% precision gives a measure of the percentage of the dataset that can be correctly recognized with no false positive matches. Ideal performance is achieved when precision and recall are equal to 100% indicating all potential matches have been reported and all reported matches are correct. The AUC metric is the integral of the Precision-Recall curve and gives another method to compare performance of different implementations, alleviating the potential extreme sensitivity of the maximum recall at 100% precision metric. Maximum performance of the AUC metric occurs at 100%, with the AUC metric giving an indication of precision over all recall levels.

For this work, true and false positives were labeled using a tolerance of 5 frames, consistent with other research in this area. Ground truth maps for the datasets were calculated by manually labeling frame correspondences between each traverse.

## V. RESULTS

In this section, we present the results from utilizing the Déjà vu algorithm testing on three different datasets.

### A. Aerial Dataset

Figures 8 and 9 present the precision-recall graph and the coverage plot at 100% recall for the Aerial Dataset. It can be seen that the Déjà vu algorithm achieves a 20.64% recall at 100% precision and achieves 95.9% precision at 100% recall. This indicates that the even through the Déjà vu algorithm does not store globally unique features, storing only repeating features within the environment, the algorithm is able to correctly localize and is able correctly recognize locations 95.9% of the time. The coverage plot is further evidence of the ability of the proposed algorithm to correctly localize within an environment utilizing only repeating features.

### B. Outdoor Dataset

The Precision-Recall graph for the Outdoor Dataset is presented in Figure 10. It shows near perfect precision for the entirety of the dataset, achieving 41.43% Recall at 100% precision and 97.85% Precision at 100% recall. The Coverage plot for the Outdoor Dataset is presented in Figure 11 and illustrates the accurate encoding and storage of the dataset enabling accurate localization within the environment.

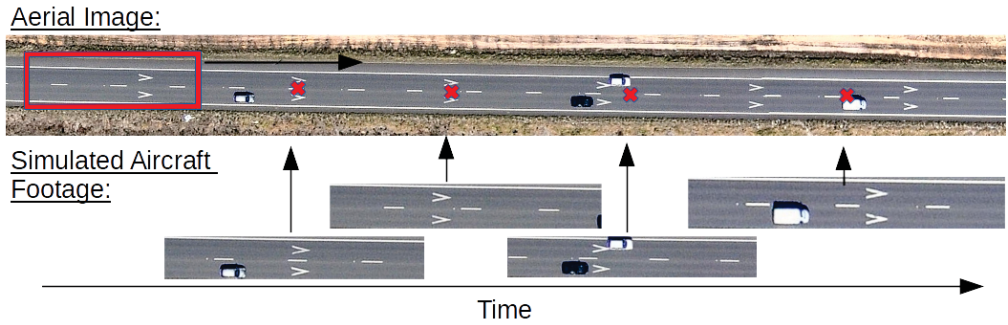

Fig. 5: Map of the Aerial Dataset with example images from the dataset. This dataset has a number of lane-markings and chevrons repeating throughout the environmentt.

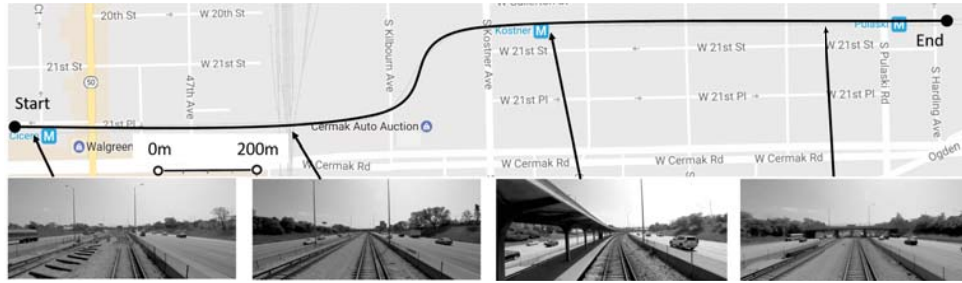

Fig. 6: Map of the Outdoor Train dataset image with a selection of images taken from along the path; note the repetition of landmarks such as light poles and supports within the train stations.

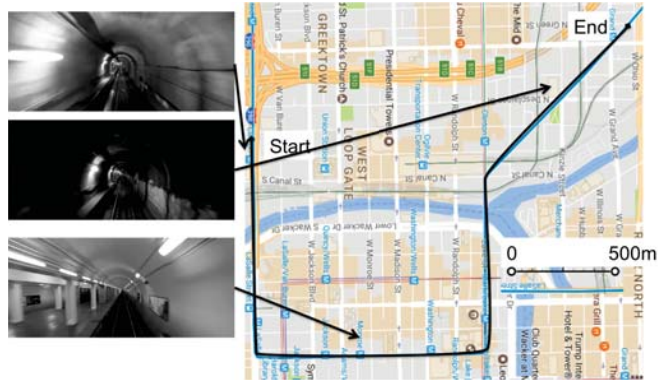

Fig. 7: Map of the Underground Train dataset image with a selection of images taken from along the path; note the repetition of landmarks such as lighting and structure within the train stations.

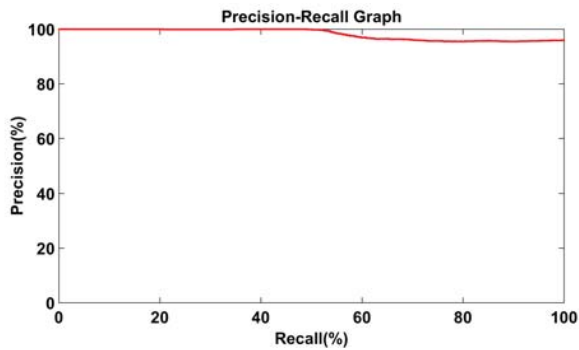

Fig. 8: Precision-Recall graph for the Aerial Dataset using the Déjà vu algorithm.

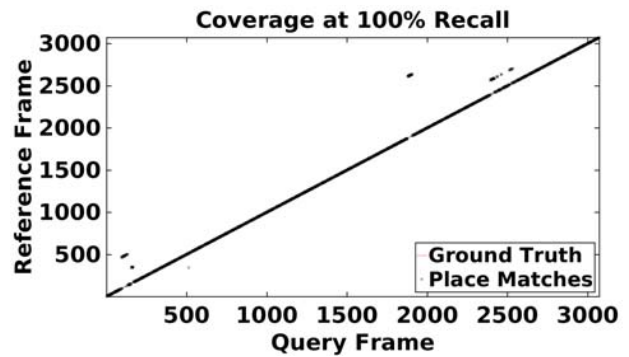

Fig. 9: Dataset Coverage graph at 100% Recall for the Aerial Dataset using the Déjà vu algorithm.

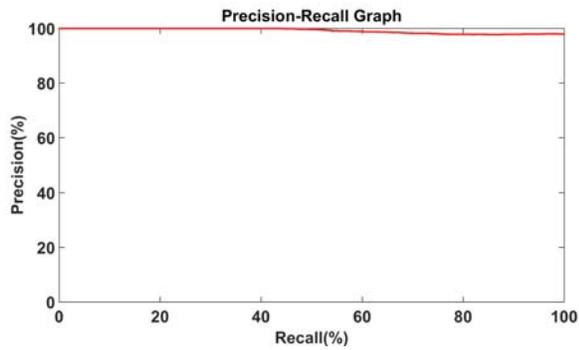

Fig. 10: Precision-Recall graph for the Outdoor Dataset using the Déjà vu algorithm.

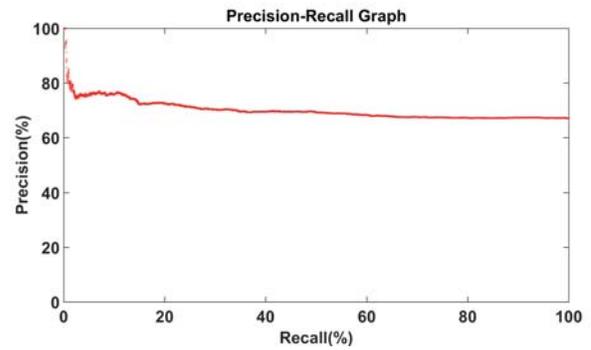

Fig. 12: Precision-Recall graph for the Underground Dataset using the Déjà vu algorithm.

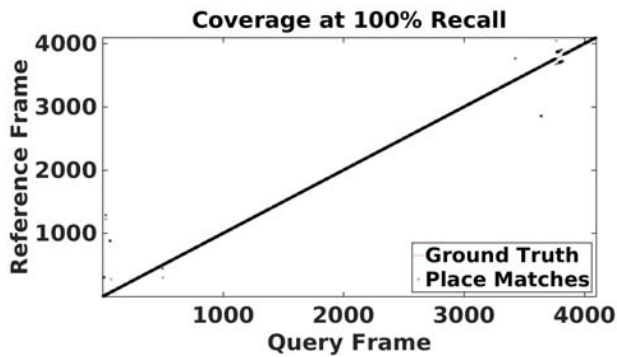

Fig. 11: Dataset Coverage graph at 100% Recall for the Outdoor Dataset using the Déjà vu algorithm.

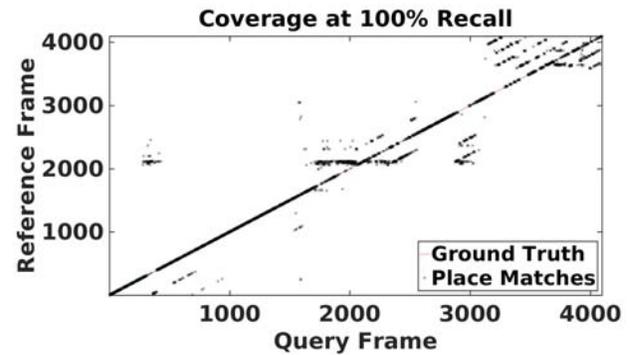

Fig. 13: Dataset Coverage graph at 100% Recall for the Underground Dataset using the Déjà vu algorithm.

*C. Underground Dataset*

The Underground Dataset was more challenging than the Aerial and the Outdoor Datasets, as seen in Figure 12, achieving only 0.32% Recall at 100% precision.

This is an example of an environment which consists of repetitive features where the fixed window spectrogram approach may not be completely suitable for identifying the start and end locations of repetitive features within this particular dataset; future work will endeavor to remove the limitations of the fixed window spectrogram for the identification of dominate features within a dataset, and the areas which they are active.

*D. Area Under The Curve*

Figure 14 shows near perfect performance for the Aerial and the Outdoor Datasets achieving scores of 98.13% and 99.1%, respectively. The Underground dataset performs poorly due to poor identification of repeating features, as discussed above, and achieves a score of 69.98%.

*E. Illustrative Results*

Figure 15 illustrates how the environment is segmented into chunks using the spectrogram and provides the resultant dominant feature periods for each segment of the dataset.

Figure 16 presents example feature templates from Chunk 1 of the Aerial Dataset. The active HOG features illustrated within the figure are active with a consistent period of 93 and 171 frames - corresponding to the translation of the dashed line marking and the chevrons respectively.

In Figure 17, we present the template activations exhibited within Chunk 1. During the first 1024 frames, there is a high number of true positive activations as the query frames move through the "chunk" which the templates were trained. Once the agent moves out of the region for which the templates are trained, the template activations become more random, forcing the algorithm to rely on other segments trained for that particular area. Fusing the template activations from Figure 17(a and b) produce reliable place estimates, as seen in Figure 17(c).

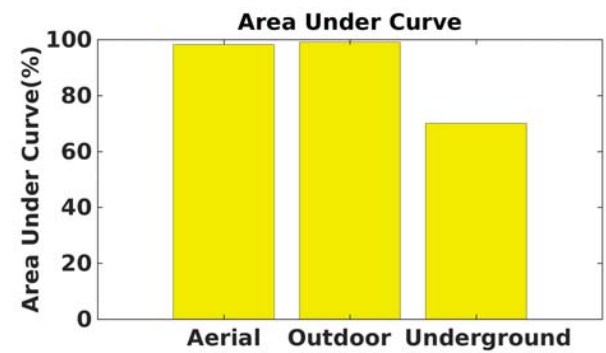

Fig. 14: Area Under Curve Graph for the Aerial, Outdoor and Underground Datasets.

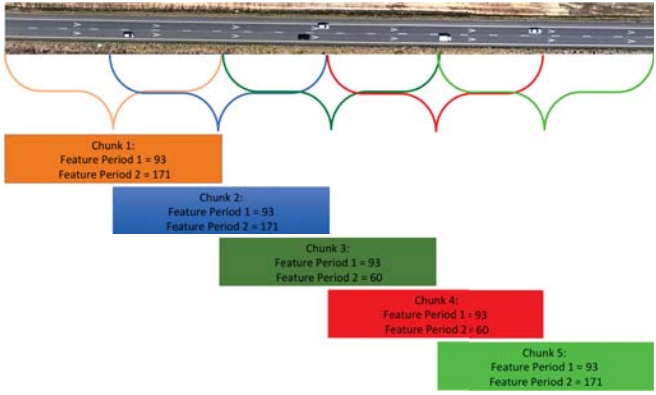

Fig. 15: Example Templates and learned periodicity for each chunk within the Aerial Dataset.

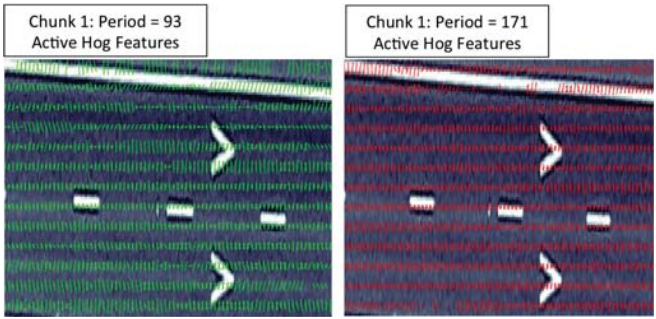

Fig. 16: Illustration of active features within Chunk 1 for the Aerial Dataset.

## VI. Discussion and Future Work

We have presented a novel method of mapping the world for the purpose of place recognition that searches for regularly-occurring features at different frequencies and uses them to reconstruct a global location estimate. The method is inspired at a high level by the repetitive neural firing patterns observed in the mammalian brain, where single neurons known as grid cells each encode a large number of places in the world at different spatial frequencies, an approach seemingly at odds with the general focus of finding visually or geometrically salient unique landmarks in many robot mapping systems. While each individual feature can only be used to disambiguate location to one of a sequence of regularly spaced locations, it is the combination of multiple features of different frequencies that enables the reconstruction of a global location. Using aerial and ground-based vision datasets we showed that the technique is capable of detecting repetitive features at different mutually supportive frequencies and successfully performing place recognition. The experiments also show that it is feasible to segment a route into chunks where each chunk is encoded by a local pair of features that repeat at different frequencies - meaning local repetition or textures can be exploited without the requirement for regular repetition throughout the entire dataset.

The experiments to date show that such repetitive features can readily be found in man-made environments like transport routes including roadways or train lines. Most promisingly, while the approach does tend to find the obvious repetitive features in the environment, it also finds repetitive features that are not intuitively obvious to a human but which nonetheless play a useful role. We are currently investigating the applicability of the approach to a wide range of other domains - both in terms of natural versus man-made environments and the platform - water-, land- or air-based. Repetitive features can be found in all environments: the question that remains to be answered is what level of environment segmentation is required to enable reliable encoding of local position by two (or more) feature frequencies. The approach described here performs global place recognition using only the current sensory snapshot. Robotic mapping and localization research has a wide variety of sequence-based localization techniques [5], [15], [16] that leverage multiple sensory snapshots to localize. Using sequences of sensory snapshots will increase the place recognition latency of this approach but may also afford further compression since it will permit a degradation in the quality of individual frame-based place matches.

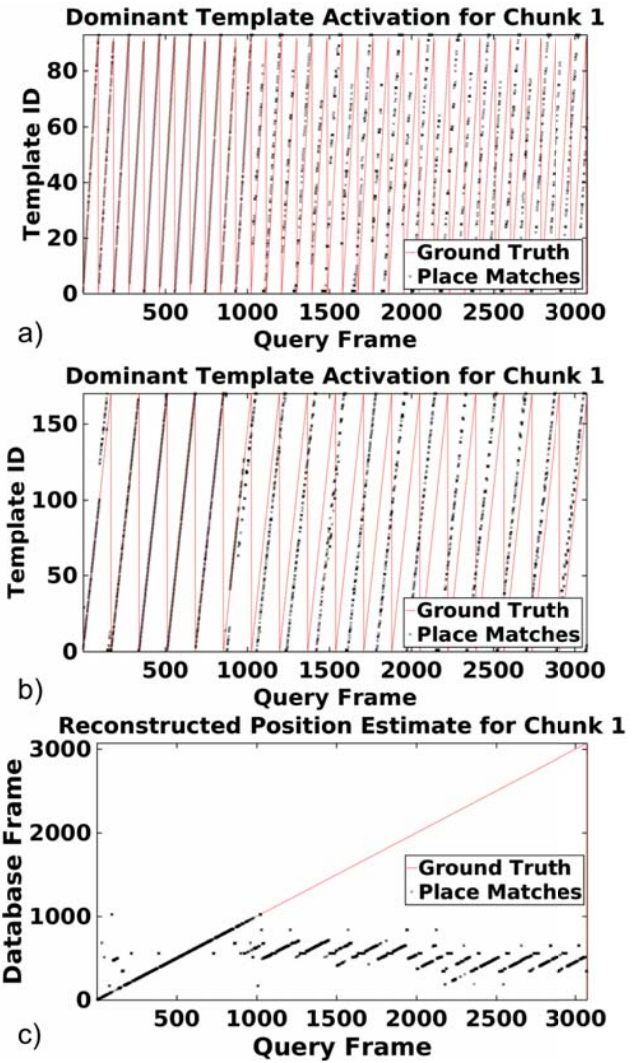

Fig. 17: Chunk 1 template activations throughout dataset with a period of 93(a) and 171 (b), with the combination place estimates(c). During the first 1024 frames, there is a high number of true positive activations as the query frames move through the "chunk" which the templates were trained. Once the agent moves out of the region for which the templates are trained, the template activations become more random, forcing the algorithm to rely on other segments trained for that particular area.

Two other areas that remain to be investigated are dimensionality and the number of different frequencies used. Grid cells in nature are known to encode two-dimensional and possibly three-dimensional environments [17] - it remains to be seen what effect scaling up the dimensionality of the approach will have. The neural encoding of space has also been observed to have many more than two spatial scales. Extending this approach to an arbitrary number of different frequencies may yield far greater encoding efficiency, since the combinatorial power grows with the number of frequencies, traded off against the difficulty of finding increasingly different but regularly repeating frequencies in any particular environment.

Finally, it will be interesting to apply this approach to other domains where large amounts of sensory data is stored and then compared with new sensory snapshots, such as in object, face and voice recognition. Storage and computational requirements are always a factor that is weighed up against algorithm accuracy and general performance: we hope that providing a novel means of efficiently encoding large amounts of data will be of utility to a wide range of real-time robotic and autonomous systems.